\documentclass[11pt,a4paper]{article}

\usepackage[utf8]{inputenc}
\usepackage[T1]{fontenc}
\usepackage{amsmath,amssymb,amsthm,mathtools}
\usepackage{booktabs}
\usepackage{graphicx}
\usepackage{hyperref}
\usepackage{cleveref}
\usepackage{xcolor}
\usepackage{microtype}
\usepackage[margin=1in]{geometry}
\usepackage{caption}
\usepackage{subcaption}
\usepackage{bm}
\usepackage{multirow}
\usepackage{enumitem}
\DeclareMathOperator*{\argmax}{arg\,max}
\newcommand{\R}{\mathbb{R}}

\newcommand{\norm}[1]{\lVert #1 \rVert}
\newcommand{\abs}[1]{\lvert #1 \rvert}

\newtheorem{definition}{Definition}

\newtheorem{prediction}{Property}

\title{Spectral Edge Dynamics of Training Trajectories:\\
Signal--Noise Geometry Across Scales}
\author{Yongzhong Xu\thanks{\texttt{abbyxu@gmail.com}.
  \quad Code: \url{https://github.com/skydancerosel/mini_gpt}}}
\date{}

\begin{document}
\maketitle

% ══════════════════════════════════════════════════════════════════════════════
\begin{abstract}
Despite hundreds of millions of parameters, transformer training trajectories
evolve within only a few coherent directions.  We introduce
\emph{Spectral Edge Dynamics} (SED) to measure this structure: rolling-window
SVD of parameter updates reveals a sharp boundary---the \emph{spectral
edge}---between coherent optimization directions and stochastic noise,
identified by the maximum consecutive singular value ratio
$\sigma_k/\sigma_{k+1}$.  Across a 51M-parameter TinyStories model (4~seeds)
and GPT-2 124M under a distribution shift, the spectral edge exhibits a
universal three-phase pattern (rise, plateau, collapse), signal rank adjusts
with task complexity ($k^* = 2$ at 51M, $k^* = 3$ at 124M), and the
directional coupling between spectral geometry and validation loss reverses
with window size---a \emph{lag flip} reflecting the timescale of trajectory
integration.  Johnson--Lindenstrauss projection to $d = 10W$ dimensions
(e.g., $d = 100$ for $W = 10$) preserves the spectral gap within 5.7\%,
making the framework applicable to models of arbitrary size.  In companion
work, the same spectral geometry provides early-warning signals of
grokking---predicting generalization 600--1{,}700 steps before it occurs
across modular arithmetic, Dyck languages, and the SCAN benchmark.
\end{abstract}

% ══════════════════════════════════════════════════════════════════════════════
\section{Introduction}\label{sec:intro}

A growing body of evidence shows that neural network training, despite
operating in spaces with hundreds of millions of dimensions, concentrates its
movement along a small number of coherent directions
\cite{gurari2018gradient,li2018measuring}.  AdamW produces a dominant backbone
capturing 60--80\% of parameter displacement \cite{xu2026optimizer}, and
spectral gap collapse in training trajectories precedes grokking by
600--1{,}700 steps across algorithmic benchmarks
\cite{xu2026execution,xu2026integrability,xu2026multitask,xu2026earlywarning}.

But \emph{where exactly is the boundary} between the signal subspace and the
noise bulk, and how does it evolve during training?  We introduce
\emph{Spectral Edge Dynamics} (SED) to answer this question.  The key object is
the singular value spectrum of a rolling window of parameter deltas, which
separates into \emph{signal} directions (coherent optimization movement) and
\emph{noise} (stochastic gradient fluctuation).  The boundary between
these---the ``spectral edge''---exhibits rich structure that we characterize
systematically across two transformer scales.

Our main findings are:

\begin{enumerate}[nosep]
\item \textbf{Signal--noise boundary of training trajectories.}
  The singular value spectrum of rolling parameter updates separates into
  a small number of coherent optimization directions and a stochastic noise
  bulk.  The boundary between them---the spectral edge---can be detected via
  the maximum consecutive singular value ratio (\Cref{sec:framework}).

\item \textbf{Cross-scale spectral geometry.}
  Across two transformer scales (51M and 124M parameters), the spectral edge
  exhibits a consistent three-phase dynamic (rise, plateau, collapse) and a
  small effective signal rank ($k^* = 2$--$3$), indicating that training
  trajectories evolve within a low-dimensional manifold
  (\Cref{sec:tinystories,sec:gpt2}).

\item \textbf{Window-dependent temporal structure.}
  The temporal relationship between the spectral edge and validation loss
  flips sign as the rolling window increases, revealing a timescale-dependent
  coupling between trajectory geometry and optimization progress
  (\Cref{sec:tinystories,sec:gpt2}).

\item \textbf{Distribution shift sensitivity.}
  The spectral edge detects a FineWeb~$\to$~OpenWebText distribution shift
  within $\pm 200$ training steps without explicit supervision
  (\Cref{sec:gpt2}).

\item \textbf{Scalable measurement.}
  Johnson--Lindenstrauss projection to dimension $d = 10W$
  (e.g., $d = 100$ for window size $W = 10$) preserves the spectral gap
  within 5--7\% error, enabling trajectory analysis for arbitrarily large
  models (\Cref{sec:scaling}).
\end{enumerate}

The theoretical foundations connecting the BBP phase transition
\cite{baik2005phase}, Benaych-Georges--Nadakuditi eigenvector alignment
\cite{bgn2011}, and Davis--Kahan subspace perturbation \cite{davis1970rotation}
to the loss--gap coupling are developed in a companion paper.  Here we focus on
the empirical characterization of spectral geometry across scales.

% ══════════════════════════════════════════════════════════════════════════════
\section{Framework}\label{sec:framework}

\subsection{Trajectory Matrix and Gram Matrix Trick}

Consider a neural network with parameters $\bm{\theta}_t \in \R^p$ trained by
an optimizer (SGD, Adam, AdamW).  At each checkpoint~$t$ we compute the
\emph{parameter delta}:
\begin{equation}\label{eq:delta}
  \bm{\delta}_t = \bm{\theta}_{t+1} - \bm{\theta}_t \in \R^p.
\end{equation}

\begin{definition}[Trajectory Matrix]\label{def:traj}
Given a rolling window of $W$ consecutive deltas starting at index~$t_0$, the
\emph{trajectory matrix} is:
\[
  \bm{X}(t_0) =
  \begin{pmatrix}
    \bm{\delta}_{t_0}^\top \\ \vdots \\ \bm{\delta}_{t_0+W-1}^\top
  \end{pmatrix}
  \in \R^{W \times p}.
\]
\end{definition}

In practice, $p \gg W$ (e.g., $p = 124\text{M}$, $W = 10$--$30$).  We
therefore work with the \emph{Gram matrix}:

\begin{definition}[Gram Matrix]\label{def:gram}
The Gram matrix is $\bm{G}(t_0) = \bm{X}\bm{X}^\top \in \R^{W \times W}$,
with entries $G_{ij} = \langle \bm{\delta}_{t_0+i-1},\,
\bm{\delta}_{t_0+j-1} \rangle$.  Its eigendecomposition
$\bm{G} = \bm{U}\bm{\Lambda}\bm{U}^\top$ yields eigenvalues
$\lambda_1 \geq \cdots \geq \lambda_W \geq 0$, with singular values
$\sigma_k = \sqrt{\lambda_k}$.
\end{definition}

The Gram matrix and the covariance matrix $\bm{X}^\top\bm{X}$ share the same
non-zero eigenvalues, so eigendecomposing the $W \times W$ Gram matrix
(complexity $O(W^3)$ after the dot products) gives exactly the same spectrum as
the intractable $p \times p$ covariance.

\paragraph{Precomputed dot-product matrix.}
For experiments testing multiple values of~$W$, we precompute the full
$N \times N$ pairwise dot-product matrix $D_{ij} = \langle \bm{\delta}_i,\,
\bm{\delta}_j \rangle$ once (complexity $O(N^2 p)$), then extract $W \times W$
submatrices for each rolling window.  This avoids redundant $p$-dimensional dot
products when sweeping over~$W$.

\subsection{Signal Rank and Spectral Gap}

\begin{definition}[Spectral Edge Ratio]\label{def:ratio}
The $k$-th \emph{consecutive singular value ratio} is:
\[
  r_k(t) = \frac{\sigma_k(t)}{\sigma_{k+1}(t)}, \qquad k = 1, \ldots, W-1.
\]
\end{definition}

\begin{definition}[Signal Rank]\label{def:kstar}
The \emph{empirical signal rank} is:
\[
  k^*(t) = \argmax_{1 \leq k \leq W-1} r_k(t).
\]
This identifies the signal--noise boundary: the first $k^*$ singular values
correspond to coherent optimization directions, and the remaining $W - k^*$
are dominated by stochastic noise.
\end{definition}

\begin{definition}[Spectral Gap]\label{def:gap}
The \emph{gap ratio} at time~$t$ is $r_{k^*}(t)$, the maximum consecutive
ratio.  We also define the \emph{edge strength} as
$r_{k^*}(t) - 1$, which measures how far the boundary is from the noise floor
(where all ratios $\approx 1$).
\end{definition}

The framework connects to random matrix theory through the BBP phase
transition \cite{baik2005phase}: in a spiked Wishart model
$\bm{X} = \bm{S} + \bm{N}$, the $k$-th signal singular value separates from
the noise bulk if and only if it exceeds a critical threshold determined by the
noise variance and aspect ratio $\gamma = p/W$.  For our extreme aspect ratio
($\gamma \sim 10^6$), this threshold is very low, so even weak signals are
detectable---but the \emph{ratio} between adjacent signal eigenvalues still
carries information about the relative strength of different learning
directions.

\subsection{Cross-Correlation Analysis}

To quantify the temporal coupling between the spectral edge ratio
$r_{k^*}(t)$ and validation loss $L_{\text{val}}(t)$, we compute the
detrended cross-correlation:

\begin{enumerate}[nosep]
\item Fit and subtract a cubic polynomial from each time series (detrending).
\item Compute the Pearson correlation between $r_{k^*}(t - \ell)$ and the
  detrended val-loss at multiple lags $\ell \in [-L, L]$.
\item The lag $\ell^*$ at which $\abs{r(\ell)}$ is maximized determines the
  temporal relationship: $\ell^* > 0$ means the spectral edge \emph{leads}
  val-loss; $\ell^* < 0$ means val-loss leads.
\end{enumerate}

A key geometric property (\Cref{pred:lagflip}) is that increasing~$W$
shifts $\ell^*$ from negative to positive, because larger windows integrate
more trajectory history and thus average over longer timescales relative to the
instantaneous val-loss.

\subsection{Geometric Properties}\label{sec:predictions}

The SED framework identifies the following geometric properties of training
trajectories, each empirically testable:

\begin{prediction}[BBP Signal Detection]\label{pred:bbp}
The spectral edge ratio $r_{k^*}$ exceeds 1.0 by a statistically significant
margin, and the noise eigenvalues' coefficient of variation exceeds the
Marchenko--Pastur prediction by $O(100\times)$ due to structured (non-isotropic)
SGD noise.
\end{prediction}

\begin{prediction}[Window-Dependent Lag Structure]\label{pred:lagflip}
The peak cross-correlation lag between the spectral edge ratio and val-loss
shifts from negative (val-loss leads) at small~$W$ to positive (spectral edge
leads) at large~$W$, reflecting the timescale of trajectory integration.
\end{prediction}

\begin{prediction}[JL Preservation]\label{pred:jl}
For projection dimension $d \geq 10W$, the spectral gap computed from
Johnson--Lindenstrauss projected deltas matches the full-dimensional value
within 10\% relative error.
\end{prediction}

\begin{prediction}[Spectral Sensitivity to Distribution Shift]\label{pred:shift}
A distribution shift produces a detectable change in the spectral edge ratio
and signal rank within $O(W)$ steps of the shift point.
\end{prediction}

\begin{prediction}[Cross-Seed Universality]\label{pred:universal}
The three-phase pattern (rise, plateau, collapse) of the spectral edge ratio is
universal across random seeds, with collapse timing varying by $\leq 5\%$ of
total training steps.
\end{prediction}

\begin{prediction}[Per-Layer Uniformity]\label{pred:perlayer}
The signal rank $k_l^*$ is approximately constant across transformer layers for
a given training stage, indicating that the effective dimensionality is a
global property of the optimization state rather than a layer-specific one.
\end{prediction}

\begin{figure}[t]
\centering
\includegraphics[width=\textwidth]{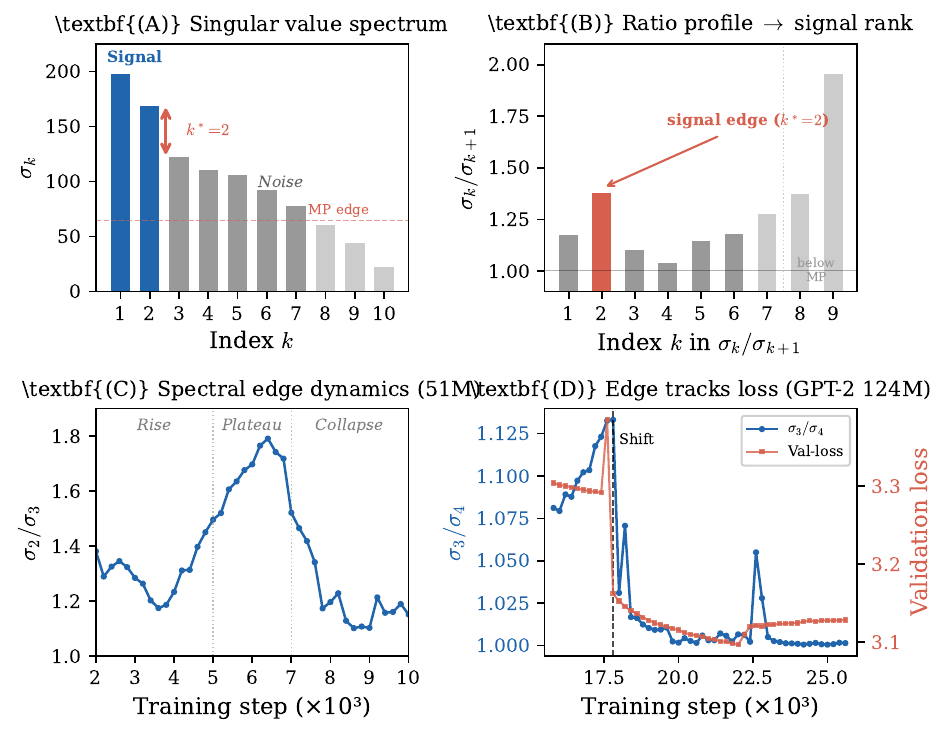}
\caption{\textbf{Overview of Spectral Edge Dynamics.}
\textbf{(A)}~Singular value spectrum of the trajectory matrix $\Delta$
(TinyStories 51M, step~2000).  The first $k^* = 2$ modes (blue) carry
the training signal; remaining modes form the noise bulk (grey).
The Marchenko--Pastur edge (dashed) separates signal from noise.
\textbf{(B)}~Consecutive ratio profile $\sigma_k/\sigma_{k+1}$: the signal
rank $k^*$ is the index of the maximum ratio above the MP edge.
\textbf{(C)}~Tracking $\sigma_2/\sigma_3$ across training reveals a
universal Rise--Plateau--Collapse pattern (TinyStories, seed~42).
\textbf{(D)}~At GPT-2 124M scale ($k^* = 3$), the spectral edge ratio
$\sigma_3/\sigma_4$ co-moves with validation loss, declining sharply at
the distribution shift (dashed line).}
\label{fig:overview}
\end{figure}

% ══════════════════════════════════════════════════════════════════════════════
\section{Experiment 1: TinyStories (51M Parameters)}\label{sec:tinystories}

\subsection{Setup}

We train a transformer language model on the TinyStories dataset
\cite{eldan2023tinystories} with 4~random seeds (42, 123, 149, 256).

\begin{center}
\renewcommand{\arraystretch}{1.1}
\begin{tabular}{ll}
\toprule
\textbf{Hyperparameter} & \textbf{Value} \\
\midrule
Architecture    & 8 layers, $d = 512$, $h = 16$, $d_{\text{ff}} = 2048$ \\
Parameters      & 51M (parameter vector: $p = 163{,}150{,}848$) \\
Optimizer       & AdamW, $\text{wd} = 0.5$, $\text{lr} = 0.001$ \\
Training        & 10{,}000 steps \\
Checkpoints     & Every 200 steps $\to$ 51 checkpoints, 50 deltas per seed \\
Window          & $W = 10$ (primary), sweep $W \in \{10, 15, 20, 25\}$ \\
\bottomrule
\end{tabular}
\end{center}

\subsection{Signal Detection and Noise Structure (Property~\ref{pred:bbp})}

Under the isotropic noise model, the Marchenko--Pastur law predicts
eigenvalue coefficient of variation (CV) of approximately $0.001$ for our
aspect ratio $\gamma = p/W = 1.6 \times 10^7$.  The observed noise CV is
$0.688$ (seed~42, step~2000), a factor of $229\times$--$576\times$ above the MP
prediction across all windows and seeds.  This confirms that the noise covariance
is heavily structured by the Adam preconditioner, as expected.

The signal rank detected by the maximum ratio method gives $k^* = 2$ across
all 4~seeds: the peak ratio is $\sigma_2/\sigma_3 = 1.747 \pm 0.120$
(cross-seed mean $\pm$ std), corresponding to two dominant learning directions
(drift and oscillation).  PC2 is above the 95th-percentile noise null in 100\%
of windows for all seeds.

\subsection{Three-Phase Spectral Pattern (Property~\ref{pred:universal})}

The spectral edge ratio $\sigma_2/\sigma_3$ exhibits a universal three-phase
pattern across all 4~seeds:

\begin{enumerate}[nosep]
\item \textbf{Rise} (steps ${\sim}1000$--$5000$): $\sigma_2/\sigma_3$
  increases from ${\sim}1.2$ to its peak value, concurrent with rapid val-loss
  improvement. 70--80\% of total val-loss reduction occurs in this phase.
\item \textbf{Plateau} (steps ${\sim}5000$--$7000$): The ratio remains elevated
  near its peak ($1.58$--$1.91$ depending on seed).
\item \textbf{Collapse} (steps ${\sim}7000$--$9000$): The ratio drops below
  1.20, indicating the second signal direction is merging with noise.  Val-loss
  stabilizes $0$--$200$ steps later (simultaneous at $W = 10$ resolution).
\end{enumerate}

\begin{figure}[t]
\centering
\includegraphics[width=0.85\linewidth]{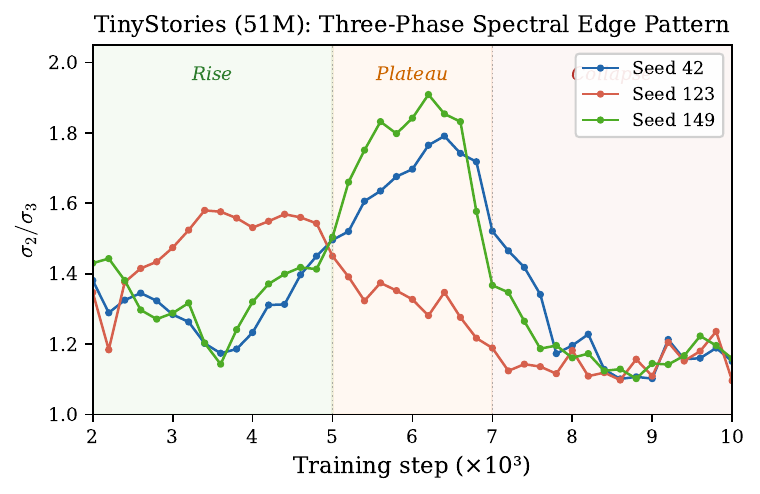}
\caption{\textbf{Three-phase spectral edge pattern across TinyStories seeds.}
The spectral edge ratio $\sigma_2/\sigma_3$ rises during rapid learning,
plateaus near its peak, and collapses as the second signal direction merges
with the noise bulk.  Collapse timing varies by $\leq 4\%$ of total training
across seeds.}
\label{fig:tinystories}
\end{figure}

The collapse occurs at step $7500 \pm 300$ across seeds (4\% of training),
confirming Property~\ref{pred:universal}.  \Cref{fig:tinystories} shows the
three-phase pattern across three random seeds.

\subsection{Phase-Specific Correlations}

Global cross-correlations can mask the true relationship due to phase
cancellation: the rise and collapse phases have \emph{opposite} correlation
signs, which cancel in a global measurement.  We therefore analyze each phase
separately.

In the collapse phase, the correlation between $\sigma_2/\sigma_3$ and
val-loss is strongly positive across all 4~seeds:

\begin{center}
\renewcommand{\arraystretch}{1.1}
\begin{tabular}{lccc}
\toprule
\textbf{Method} & \textbf{Mean $r$} & \textbf{Std $r$} & \textbf{Seeds} \\
\midrule
Peak-based segmentation & $0.864$ & $0.059$ & $[0.93, 0.88, 0.88, 0.77]$ \\
Derivative-based        & $0.937$ & $0.019$ & $[0.96, 0.93, 0.94, 0.91]$ \\
Threshold-based         & $0.643$ & $0.060$ & $[0.57, 0.64, 0.74, 0.62]$ \\
\bottomrule
\end{tabular}
\end{center}

The derivative-based segmentation, which identifies phase boundaries from
sign changes in $d(\sigma_2/\sigma_3)/dt$, yields the most robust result:
$r = 0.937 \pm 0.019$ across all 4~seeds.

\subsection{Lag Flip with Window Size (Property~\ref{pred:lagflip})}

\begin{center}
\renewcommand{\arraystretch}{1.1}
\begin{tabular}{cccc}
\toprule
$W$ & \textbf{Peak lag} & \textbf{Peak $r$} & \textbf{Interpretation} \\
\midrule
10 & $-1$ & $-0.738$ & Val-loss leads (1 step) \\
15 & $-2$ & $-0.089$ & Transition zone \\
20 & $+1$ & $-0.547$ & Spectral gap leads! \\
25 & $+4$ & $-0.150$ & Gap leads (weak signal) \\
\bottomrule
\end{tabular}
\end{center}

The lag flips sign between $W = 15$ and $W = 20$, confirming
Property~\ref{pred:lagflip}: larger windows integrate more trajectory history,
enabling the spectral gap to become a \emph{leading} indicator of val-loss.

\subsection{Comparison with Drift Speed}

We compare the coupling strength of $\sigma_2/\sigma_3$ against drift speed
$\norm{\bar{\bm{\delta}}}$ at $W = 10$:

\begin{center}
\renewcommand{\arraystretch}{1.1}
\begin{tabular}{lcccc}
\toprule
\textbf{Seed} & $\abs{r}_{\sigma_2/\sigma_3}$ & \textbf{Lag} &
  $\abs{r}_{\text{drift}}$ & \textbf{Lag} \\
\midrule
42  & $0.738$ & $-1$ & $0.711$ & $+1$ \\
123 & $0.497$ & $-1$ & $0.366$ & $-3$ \\
149 & $0.296$ & $-3$ & $0.666$ & $\phantom{-}0$ \\
256 & $0.185$ & $-1$ & $0.273$ & $\phantom{-}0$ \\
\bottomrule
\end{tabular}
\end{center}

The spectral edge ratio and drift speed are complementary: $\sigma_2/\sigma_3$
wins for seeds 42 and 123, while drift speed wins for seeds 149 and 256.
Neither dominates uniformly, suggesting that spectral edge ratio and drift
speed capture complementary aspects of the trajectory geometry.

\subsection{Johnson--Lindenstrauss Projection (Property~\ref{pred:jl})}

We test whether random projection preserves the spectral gap using streaming
JL projection (3~projection seeds) at dimensions $d \in \{50, 100, 200, 500,
1000\}$:

\begin{center}
\renewcommand{\arraystretch}{1.1}
\begin{tabular}{cccc}
\toprule
$d$ & $d/W$ & \textbf{Mean rel.\ error ($\sigma_2/\sigma_3$)}
  & \textbf{Max rel.\ error} \\
\midrule
50   & 5   & 9.6\% & 22.4\% \\
100  & 10  & 7.4\% & 20.2\% \\
200  & 20  & 5.4\% & 17.0\% \\
500  & 50  & 3.0\% & 10.5\% \\
1000 & 100 & 2.5\% & 6.4\% \\
\bottomrule
\end{tabular}
\end{center}

At $d = 100$ ($10W$), the mean relative error is 7.4\%, confirming
Property~\ref{pred:jl}.  This means that for a 7B or 70B model, one needs
only to store $d \sim 100$--$200$ projected coordinates per checkpoint rather
than the full parameter vector.

% ══════════════════════════════════════════════════════════════════════════════
\section{Experiment 2: GPT-2 124M (Distribution Shift)}\label{sec:gpt2}

\subsection{Setup}

We study GPT-2 124M \cite{radford2019language} through a pretraining and
fine-tuning pipeline with an explicit distribution shift:

\begin{center}
\renewcommand{\arraystretch}{1.1}
\begin{tabular}{ll}
\toprule
\textbf{Phase} & \textbf{Details} \\
\midrule
Architecture & GPT-2 124M: 12 layers, $d = 768$, 12 heads \\
Parameters   & $p = 124{,}412{,}160$ (excl.\ causal masks, tied lm\_head) \\
Pretraining  & FineWeb 10B, steps 0--17{,}600, cosine LR, H100 \\
Fine-tuning  & OpenWebText, steps 17{,}800--25{,}600, const.\ LR $3\times 10^{-5}$, M4 Max \\
Shift point  & Combined step 17{,}800 \\
Overfit onset& Combined step $\sim$22{,}200 \\
Checkpoints  & 60 combined (steps 13{,}800--25{,}600, every 200) \\
\bottomrule
\end{tabular}
\end{center}

Val-loss trajectory: $3.333$ (step 13{,}800) $\to$ $3.291$ (17{,}600)
$\to$ $3.162$ (17{,}800, shift!) $\to$ $3.095$ (22{,}000, minimum)
$\to$ $3.128$ (25{,}600, overfitting).

\paragraph{Key normalization.}
FineWeb checkpoints use the \texttt{\_orig\_mod.} prefix from
\texttt{torch.compile}; OWT checkpoints do not.  We strip this prefix for
consistency.  We also skip \texttt{attn.bias} (causal mask, not a learned
parameter) and \texttt{lm\_head.weight} (tied to \texttt{wte.weight}).

\subsection{Signal Rank and Spectral Structure}

Unlike TinyStories where $k^* = 2$ and the key edge is $\sigma_2/\sigma_3$,
GPT-2 124M has $k^* = 3$ and the key edge is $\sigma_3/\sigma_4$.  We compute
\emph{all} consecutive ratios $\sigma_k/\sigma_{k+1}$ for $k = 1, \ldots, W-1$
and rank them by cross-correlation with val-loss:

\begin{center}
\renewcommand{\arraystretch}{1.1}
\begin{tabular}{lccc}
\toprule
\textbf{Ratio} & $\abs{r}_{\text{peak}}$ & \textbf{Peak lag} & \textbf{Note} \\
\midrule
$\sigma_3/\sigma_4$ & $0.870$ & $-1$ & \textbf{Best} \\
$\sigma_4/\sigma_5$ & $0.858$ & $-2$ & Close second \\
$\sigma_5/\sigma_6$ & $0.793$ & $-5$ & \\
drift speed         & $0.603$ & $-3$ & \\
$\sigma_1/\sigma_2$ & $0.565$ & $-2$ & \\
$\sigma_2/\sigma_3$ & $0.211$ & $-2$ & Weakest \\
\bottomrule
\end{tabular}
\end{center}

The correct spectral edge automatically emerges as the ratio most strongly
coupled to training dynamics.
Critically, $\sigma_2/\sigma_3$---which was the key ratio at 51M
scale---is the \emph{weakest} correlate at 124M, demonstrating that the
signal rank $k^*$ adjusts automatically with model and task complexity.

The noise eigenvalue CV ranges from 0.26 to 0.50, compared to the MP
prediction of 0.001---a factor of $229\times$--$576\times$
(Property~\ref{pred:bbp}).  At this extreme aspect ratio ($\gamma \sim 10^6$),
the MP-based signal rank detector gives $k^*_{\text{MP}} = 7$ (all eigenvalues
above the MP edge), confirming that the noise covariance is heavily structured
by the Adam preconditioner.  The ratio-based detector correctly identifies
$k^* \in \{1, 2\}$ depending on training phase, tracking the transition from
multi-directional to single-directional dynamics.

\begin{figure}[t]
\centering
\includegraphics[width=0.85\linewidth]{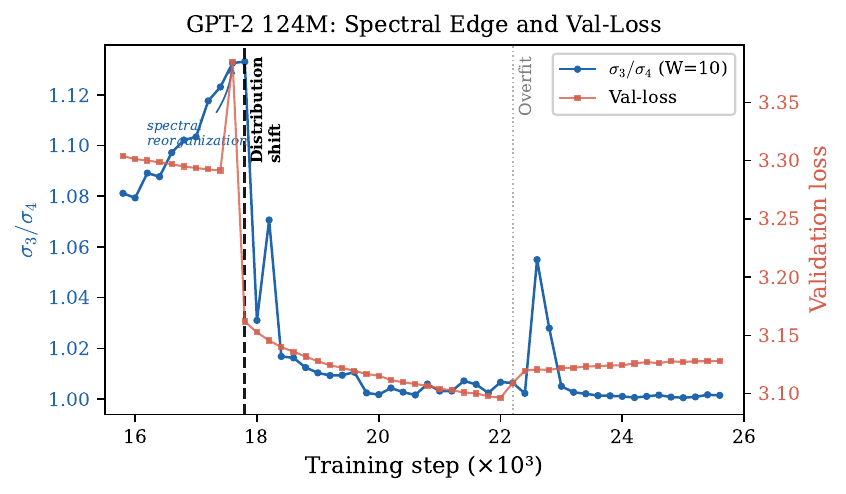}
\caption{\textbf{GPT-2 124M: Spectral edge ratio and validation loss.}
The spectral edge ratio $\sigma_3/\sigma_4$ (blue, left axis) is tightly
coupled to val-loss (red, right axis).  The distribution shift at step
17{,}800 produces a sharp spectral reorganization: the edge ratio drops
abruptly, recovers during rapid improvement, then flattens as overfitting
begins around step 22{,}200.}
\label{fig:gpt2edge}
\end{figure}

Among 6~spectral observables at $W = 10$, the edge ratio
$\sigma_3/\sigma_4$ is the most tightly coupled to training dynamics
($\abs{r} = 0.870$, lag~$= -1$), exceeding total variance ($0.620$),
drift speed ($0.603$), and $k_{95}$ ($0.439$).
\Cref{fig:gpt2edge} shows the spectral edge ratio alongside val-loss
across the distribution shift.

\subsection{Distribution Shift Response (Property~\ref{pred:shift})}

The spectral edge detects the FineWeb~$\to$~OpenWebText distribution shift
at step 17{,}800 without explicit supervision:

\begin{center}
\renewcommand{\arraystretch}{1.1}
\begin{tabular}{lcc}
\toprule
\textbf{Method} & \textbf{Detected step} & \textbf{Error} \\
\midrule
Max derivative of $\sigma_3/\sigma_4$ & 18{,}000 & $+200$ steps \\
CUSUM of $\sigma_3/\sigma_4$          & 18{,}200 & $+400$ steps \\
Two-sample $t$-test ($t = 14.9$)      & 18{,}000 & $+200$ steps \\
\bottomrule
\end{tabular}
\end{center}

All methods detect the shift within 200--400 steps ($\leq 2W$) of the true
shift point.  Split-half reliability confirms the robustness of these
measurements: the edge position matches between odd-indexed and even-indexed
half-windows in 42 of 50 windows (84\%), with mismatches concentrated in the
transition zone immediately after the shift (steps 18{,}200--19{,}400).  The
ratio profile correlation has median $> 0.99$ outside the transition zone.

\subsection{Temporal Structure: Lag Flip and Granger Causality}\label{sec:granger}

\paragraph{Lag flip (Property~\ref{pred:lagflip}).}

\begin{center}
\renewcommand{\arraystretch}{1.1}
\begin{tabular}{lcccc}
\toprule
$W$ & \textbf{Best ratio} & $\abs{r}$ & \textbf{Lag} & \textbf{Interpretation} \\
\midrule
10 & $\sigma_3/\sigma_4$ & $0.870$ & $-1$ & Val-loss leads \\
20 & $\sigma_1/\sigma_2$ & $0.797$ & $\phantom{+}0$ & Simultaneous \\
30 & $\sigma_1/\sigma_2$ & $0.882$ & $+2$ & Spectral gap leads \\
\bottomrule
\end{tabular}
\end{center}

The lag flips sign from $-1$ to $+2$ as $W$ increases from 10 to 30:
larger windows integrate more trajectory history, enabling the spectral gap
to become a \emph{leading} indicator of val-loss.

\paragraph{Granger causality.}
The lag flip implies that the directional Granger coupling should also reverse.
At $W = 10$, the dominant direction is val-loss $\to$ $\sigma_3/\sigma_4$
($F = 54.4$, $p = 1.7 \times 10^{-12}$), while the reverse is marginal
($F = 3.3$, $p = 0.046$).  At $W = 20$, the asymmetry \emph{flips}:

\begin{center}
\renewcommand{\arraystretch}{1.1}
\begin{tabular}{llccc}
\toprule
$W$ & \textbf{Direction} & $F$ & $p$ & $\Delta R^2$ \\
\midrule
\multirow{2}{*}{10} & $\sigma_3/\sigma_4 \to \text{VL}$ & 3.32 & 0.046 & 13.4\% \\
                    & $\text{VL} \to \sigma_3/\sigma_4$ & 54.4 & $1.7 \times 10^{-12}$ & 71.7\% \\
\midrule
\multirow{2}{*}{20} & $\sigma_3/\sigma_4 \to \text{VL}$ & 11.2 & \textbf{0.002} & 23.8\% \\
                    & $\text{VL} \to \sigma_3/\sigma_4$ & 2.22 & 0.093 (ns) & 24.8\% \\
\midrule
\multirow{2}{*}{30} & $\sigma_3/\sigma_4 \to \text{VL}$ & 3.40 & \textbf{0.032} & 44.4\% \\
                    & $\text{VL} \to \sigma_3/\sigma_4$ & 10.4 & 0.003 & 28.6\% \\
\bottomrule
\end{tabular}
\end{center}

At $W = 20$, $\sigma_3/\sigma_4$ Granger-causes val-loss ($p = 0.002$)
while the reverse direction becomes non-significant ($p = 0.093$).

\subsection{Per-Layer Uniformity (Property~\ref{pred:perlayer})}

We group the 124M parameters into 8~layer groups and compute per-group spectral
edge ratios.  The cross-correlation of each group's $\sigma_3/\sigma_4$ with
val-loss is:

\begin{center}
\renewcommand{\arraystretch}{1.1}
\begin{tabular}{lcccc}
\toprule
\textbf{Layer group} & $p_l$ & $\abs{r}_{\text{peak}}$ & \textbf{Lag}
  & $\abs{r}_{\sigma_1/\sigma_2}$ \\
\midrule
attn\_early (L0--3) & 9.4M  & $\mathbf{0.921}$ & $-1$ & $0.583$ \\
attn\_mid (L4--7)   & 9.4M  & $0.895$ & $-1$ & $0.721$ \\
mlp\_early (L0--3)  & 18.9M & $0.893$ & $-1$ & $0.706$ \\
mlp\_mid (L4--7)    & 18.9M & $0.871$ & $-1$ & $0.769$ \\
mlp\_late (L8--11)  & 18.9M & $0.848$ & $-1$ & $0.761$ \\
embedding           & 39.4M & $0.840$ & $-1$ & $0.419$ \\
attn\_late (L8--11) & 9.4M  & $0.824$ & $-1$ & $0.770$ \\
layernorm           & 38K   & $0.824$ & $-2$ & $0.333$ \\
\bottomrule
\end{tabular}
\end{center}

All layer groups show strong correlations ($\abs{r} > 0.82$) with consistent
lag~$= -1$, confirming Property~\ref{pred:perlayer}.  Early attention layers
achieve the highest correlation ($0.921$), suggesting they are most sensitive
to the distribution shift and training dynamics.

\subsection{Information Content of the Spectral Edge}

\subsubsection{Residualized Spectral Edge}

To test whether the spectral edge carries geometric information beyond
what is already captured by the loss trajectory, we regress out
val-loss from $\sigma_3/\sigma_4$ and test whether the \emph{residual}
still Granger-causes val-loss:

\begin{center}
\renewcommand{\arraystretch}{1.1}
\begin{tabular}{ccccc}
\toprule
$W$ & \textbf{Lags} & $R^2_{\text{explained}}$ & \textbf{Residual Granger $p$}
  & $\Delta R^2$ \\
\midrule
10 & 1 & 87.8\% & \textbf{0.018} & 11.8\% \\
10 & 2 & 88.7\% & \textbf{0.027} & 10.6\% \\
20 & 1 & 24.3\% & \textbf{0.007} & 19.1\% \\
20 & 2 & 30.7\% & \textbf{0.026} & 13.8\% \\
\bottomrule
\end{tabular}
\end{center}

Even at $W = 10$, where val-loss explains 88\% of the spectral edge variance,
the residual still significantly Granger-causes val-loss ($p = 0.018$).  At
$W = 20$, the effect is stronger ($p = 0.007$).  This confirms that the
spectral edge captures geometric structure in the trajectory that is not
reducible to the loss signal alone.

\subsubsection{Multivariate Granger Analysis}

Combining multiple spectral observables in a joint Granger test reveals
the information content of different geometric features:

\begin{center}
\renewcommand{\arraystretch}{1.1}
\begin{tabular}{lcccc}
\toprule
\textbf{Observable set} & $W$ & $p$ & $\Delta R^2$ & $F$ \\
\midrule
$\sigma_3/\sigma_4$ alone         & 10 & 0.046 & 13.4\% & 3.3 \\
$k_{95}$ alone                    & 10 & $6.1 \times 10^{-5}$ & 47.7\% & 8.4 \\
edge\_str $+$ $k_{95}$            & 10 & $\mathbf{1.1 \times 10^{-16}}$ & \textbf{89.5\%} & 52.6 \\
All 5 observables                  & 10 & $9.9 \times 10^{-14}$ & 94.6\% & 32.4 \\
\midrule
$\sigma_3/\sigma_4$ alone         & 20 & 0.002 & 23.8\% & 11.2 \\
$k_{95} +$ drift                  & 20 & $3.7 \times 10^{-8}$ & 79.5\% & 17.5 \\
All 5 observables                  & 20 & $5.6 \times 10^{-9}$ & 95.4\% & 24.7 \\
\bottomrule
\end{tabular}
\end{center}

The pair (edge\_strength, $k_{95}$) is particularly informative: at $W = 10$,
it explains $89.5\%$ of val-loss variance beyond the autoregressive baseline
($p = 1.1 \times 10^{-16}$, $F = 52.6$).  This demonstrates that the
spectral edge ratio and effective rank capture complementary aspects of the
trajectory geometry.

\subsubsection{Phase-Specific Sliding-Window Correlations}

\begin{figure}[t]
\centering
\includegraphics[width=0.85\linewidth]{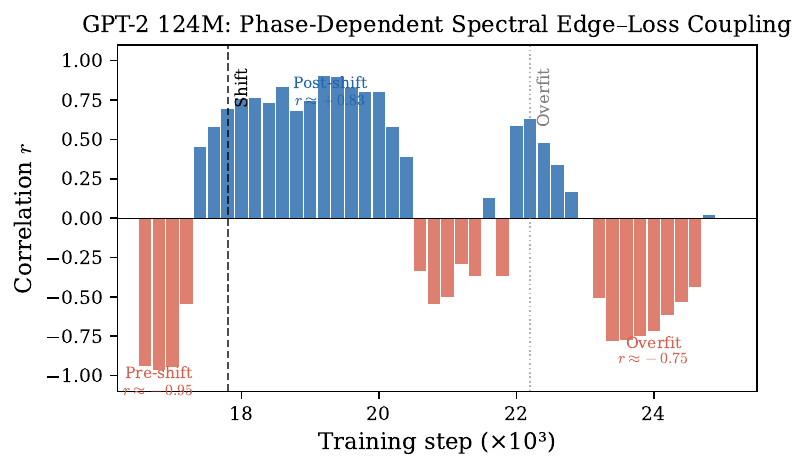}
\caption{\textbf{Phase-dependent spectral edge--loss coupling.}
Sliding-window Pearson correlation between $\sigma_3/\sigma_4$ and val-loss
(window of 7~steps, $W = 10$).  The correlation sign flips across training
phases: strongly negative during late pretraining ($r \approx -0.95$),
strongly positive after the distribution shift ($r \approx +0.83$), and
negative again during overfitting ($r \approx -0.75$).}
\label{fig:slidingcorr}
\end{figure}

Using a sliding window of 7~steps, we compute local correlations between
$\sigma_3/\sigma_4$ and val-loss throughout training
(\Cref{fig:slidingcorr}).  At $W = 10$:

\begin{itemize}[nosep]
\item \textbf{Pre-shift (steps 16{,}600--17{,}200):} $r \in [-0.94, -0.96]$
  --- near-perfect anti-correlation during late pretraining.
\item \textbf{Post-shift rapid improvement (steps 18{,}000--20{,}000):}
  $r \in [+0.70, +0.90]$ --- sign flip reflects new learning dynamics.
\item \textbf{Overfitting onset (steps 23{,}000--25{,}000):}
  $r$ reverses again to $[-0.53, -0.78]$ as training becomes unproductive.
\end{itemize}

The mean absolute sliding-window correlation is $0.58$ at $W = 10$ and $0.53$
at $W = 20$, confirming that the relationship between spectral edge and
val-loss is strong but phase-dependent---consistent with the phase cancellation
observed in TinyStories.

% ══════════════════════════════════════════════════════════════════════════════
\section{Scalability via Random Projection}\label{sec:scaling}

The SED framework requires computing $p$-dimensional dot products between
parameter deltas, where $p$ can be $10^8$--$10^{11}$ for modern language
models.  The Johnson--Lindenstrauss lemma \cite{johnson1984extensions}
guarantees that random projection $\bm{\Phi} \in \R^{d \times p}$ preserves
all pairwise distances up to factor $(1 \pm \varepsilon)$ for
$d = O(\varepsilon^{-2} \log N)$ where $N$ is the number of delta vectors.

For our streaming implementation, we use row-by-row random Gaussian projection:
each projected coordinate
$\tilde{\delta}_{t,j} = \langle \bm{\delta}_t, \bm{\phi}_j \rangle$ where
$\bm{\phi}_j$ is generated on-the-fly from a seeded RNG, avoiding storage of
the $d \times p$ projection matrix.

Our empirical results across both scales confirm that $d = 10W$ suffices:

\begin{center}
\renewcommand{\arraystretch}{1.1}
\begin{tabular}{lccccc}
\toprule
\textbf{Scale} & $p$ & $W$ & $d$ & $d/W$ & \textbf{Mean rel.\ error} \\
\midrule
TinyStories 51M  & 163M & 10 & 100 & 10 & 7.4\% \\
GPT-2 124M       & 124M & 10 & 100 & 10 & 5.7\% \\
TinyStories 51M  & 163M & 10 & 500 & 50 & 3.0\% \\
GPT-2 124M       & 124M & 10 & 500 & 50 & 3.1\% \\
\bottomrule
\end{tabular}
\end{center}

Since the projection dimension $d$ depends only on $W$ (not on $p$), the
framework scales to models of arbitrary size.  For a 70B-parameter model with
$W = 20$, one would need $d \approx 200$, requiring storage of only $200$
floats per checkpoint rather than $70 \times 10^9$.

% ══════════════════════════════════════════════════════════════════════════════
\section{Cross-Scale Comparison}\label{sec:crossscale}

\begin{center}
\renewcommand{\arraystretch}{1.1}
\begin{tabular}{lcc}
\toprule
\textbf{Property} & \textbf{TinyStories (51M)} & \textbf{GPT-2 (124M)} \\
\midrule
Signal rank $k^*$      & 2 (oscillatory)   & 3 (monotonic + shift) \\
Key spectral edge      & $\sigma_2/\sigma_3$ & $\sigma_3/\sigma_4$ \\
Edge peak value        & $1.79$            & ${\sim}1.13$ \\
Edge mean value        & $1.36$            & ${\sim}1.08$ \\
Best $\abs{r}$ ($W = 10$) & $0.738$        & $0.870$ \\
Lag flip ($W$: 10$\to$20) & $-1 \to +1$    & $-1 \to 0$ \\
Training regime        & Single dataset    & Distribution shift \\
BBP noise CV ratio     & $229\times$--$693\times$ & $229\times$--$576\times$ \\
JL error at $d = 10W$  & 7.4\%             & 5.7\% \\
\midrule
Phase pattern          & 3-phase (rise/plateau/collapse)
  & 3-phase (pre-shift/rapid-improve/overfit) \\
Phase-specific $\abs{r}$ & $0.937 \pm 0.019$ & $0.944$ (pre-shift) \\
Granger (gap$\to$VL)   & ---               & $p = 0.002$ ($W = 20$) \\
\bottomrule
\end{tabular}
\end{center}

Several patterns are robust across scales:
\begin{enumerate}[nosep]
\item The lag flip from val-loss-leading to gap-leading as $W$ increases.
\item Strong phase-specific correlations ($\abs{r} > 0.9$) despite weaker
  global correlations.
\item JL projection errors of $\leq 10\%$ at $d = 10W$.
\item Structured noise far exceeding the isotropic MP prediction.
\end{enumerate}

The key difference is the signal rank: $k^* = 2$ for TinyStories (drift +
oscillation) versus $k^* = 3$ for GPT-2 (an additional direction associated
with the distribution shift).  This is consistent with the interpretation that signal rank reflects the number
of ``active'' curvature directions in the Hessian, which increases with task
complexity.

% ══════════════════════════════════════════════════════════════════════════════
\section{Related Work}\label{sec:related}

\paragraph{Low-dimensional trajectory.}
Gur-Ari et al.~\cite{gurari2018gradient} and Li et
al.~\cite{li2018measuring} established that SGD lives in a low-dimensional
subspace.  They measure rank but do not study the \emph{boundary} between
signal and noise, do not connect to RMT, and do not make predictive claims
about loss dynamics.

\paragraph{Weight matrix spectral dynamics.}
Martin \& Mahoney \cite{martin2021implicit} and Balsells~Rodas et
al.~\cite{balsells2025sgd} study the SVD of weight matrices
$\bm{W}^{(l)}(t)$ themselves, showing power-law tails and Dyson Brownian
motion.  This concerns the structure of the \emph{model}, not the
\emph{trajectory}.  The trajectory SVD is a different mathematical object
capturing optimization dynamics rather than representational structure.

\paragraph{Edge of stability.}
Cohen et al.~\cite{cohen2021gradient} show that the largest Hessian eigenvalue
hovers at $2/\eta$ during training.  Our framework is complementary: the
Hessian eigenvalues determine the signal rank $k^*$, and the spectral gap
tracks when a Hessian direction becomes active or inactive.

\paragraph{BBP transition in ML.}
The BBP transition \cite{baik2005phase} has been applied to neural networks at
\emph{initialization} for signal propagation analysis
\cite{pennington2017nonlinear,pennington2018spectrum}.  Applying it to
\emph{training trajectories} is, to our knowledge, new.

\paragraph{Grokking and spectral geometry.}
The grokking phenomenon \cite{power2022grokking}---where delayed generalization
follows memorization---has been analyzed through mechanistic interpretability
\cite{nanda2023progress} and circuit competition \cite{merrill2023tale}.
In companion work, we have shown that spectral properties of the training
trajectory provide early-warning signals of grokking across multiple task
families.  On modular arithmetic, the spectral gap $\sigma_2/\sigma_3$
collapses at step~1{,}400, predicting grokking at step~3{,}100 with a
1{,}700-step lead time \cite{xu2026integrability}.  This extends to
Dyck grammars ($\alpha \approx 1.13$) and the SCAN compositional benchmark
($\alpha \approx 1.18$) \cite{xu2026earlywarning}, and to multi-task
settings where 42/42 single-task and 27/27 tri-task conditions all exhibit
the pattern \cite{xu2026multitask}.  An execution-manifold analysis
\cite{xu2026execution} further reveals that training evolves within a
low-dimensional integrable subspace in weight space.  The present paper
characterizes the same spectral geometry at larger scale (51M--124M),
where grokking does not occur but the geometric structure persists.

\paragraph{Optimizer-induced trajectory structure.}
Our prior work \cite{xu2026optimizer} showed that AdamW induces a dominant
low-dimensional drift backbone that is absent under SGD.  The present
paper unifies these observations into the SED framework and provides
systematic empirical characterization across scales.

% ══════════════════════════════════════════════════════════════════════════════
\section{Discussion}\label{sec:discussion}

\paragraph{Geometric structure of training trajectories.}
The experiments in this paper suggest that neural network training trajectories
possess a structured signal--noise geometry.  Rather than exploring the full
parameter space, the optimizer moves primarily along a small set of coherent
directions that correspond to active curvature modes of the loss landscape.
The spectral edge marks the boundary between these coherent learning
directions and stochastic gradient fluctuations.

When this boundary is well separated, training progresses along stable
directions aligned with the loss surface.  When the boundary collapses, the
optimizer's motion becomes less coherent, and the relationship between
trajectory geometry and loss dynamics changes qualitatively.  This perspective
provides a complementary view of optimization dynamics alongside Hessian-based
analyses such as the edge-of-stability phenomenon \cite{cohen2021gradient}.
The formal connection runs through the BBP eigenvector alignment theorem
\cite{bgn2011} and the Davis--Kahan subspace perturbation bound
\cite{davis1970rotation}, which show that subspace estimation degrades as
$O(1/g)$ where $g$ is the spectral gap.

\paragraph{Backbone manifold and transverse noise.}
These results suggest a simplified geometric model of training dynamics:
parameter updates decompose as $\bm{\delta}_t \approx \bm{s}_t + \bm{n}_t$,
where $\bm{s}_t$ lies in a low-dimensional \emph{backbone manifold} and
$\bm{n}_t$ is predominantly transverse stochastic noise.  The spectral edge
identifies the evolving dimension of this backbone manifold.  In our
experiments, $\dim(\text{backbone}) \approx k^* = 2$--$3$, which is
remarkably small compared to the parameter count ($p \sim 10^8$).

The three-phase
spectral pattern then describes the evolution of this backbone manifold: the
signal subspace grows during early learning, stabilizes during productive
training, and merges with the noise bulk as the optimizer exhausts the
low-dimensional landscape structure.  We emphasize that this is an empirical
observation suggesting a theoretical conjecture, not a proven result; the
degree to which training trajectories concentrate near such a manifold, and the
stability of the backbone under perturbation, remain open mathematical
questions.

\paragraph{Is the low signal rank an artifact of $W$?}
Since $\Delta$ has at most $W$ columns, one might suspect that finding
$k^* = 2$--$3$ reflects the small matrix rather than genuine structure.
Three lines of evidence rule this out.
\emph{First}, under the Marchenko--Pastur null at our aspect ratio
$\gamma = p/W > 10^7$, all consecutive ratios converge to~$1$;
the observed $\sigma_2/\sigma_3 = 1.75 \pm 0.12$ is many standard deviations
above this null (\S\ref{sec:tinystories}).
\emph{Second}, $k^*$ is invariant across window sizes $W \in \{10, 15, 20, 25\}$
(\S\ref{sec:tinystories}) and across model scales
($k^* = 2$ at 51M, $k^* = 3$ at 124M; \S\ref{sec:crossscale});
a window artifact would vary with~$W$.
\emph{Third}, no static null model explains the \emph{dynamics}---the
three-phase rise--plateau--collapse of the ratio, its Granger-causal coupling
to val-loss, and the sign flip of this coupling with~$W$.

\paragraph{Window size and temporal scale.}
A window of size $W$ captures $W$ consecutive deltas, spanning $200W$ training
steps in our checkpoint spacing.  At $W = 10$, the window covers 2{,}000
steps, providing a local snapshot that responds quickly to loss changes.
At $W = 20$--$30$, the window covers 4{,}000--6{,}000 steps, integrating
enough trajectory history to average out stochastic fluctuations and capture
longer-timescale geometric trends.  The lag flip with increasing $W$ is a
natural consequence of this temporal integration.

\paragraph{Structured noise and the BBP threshold.}
The isotropic noise assumption underlying the standard BBP transition fails
spectacularly in our data: the noise eigenvalue spread exceeds the
Marchenko--Pastur prediction by $200$--$600\times$.  This is expected because
Adam's second-moment preconditioner creates strongly anisotropic noise.  The
colored-noise generalization of the BBP transition \cite{bgn2012} is required
for the theoretical treatment, and the noise covariance can in principle be
estimated from the Adam state variables.

\paragraph{Connection to grokking.}
In companion work on algorithmic tasks \cite{xu2026integrability,
xu2026earlywarning, xu2026multitask}, the same spectral geometry provides
early-warning signals of grokking---a setting where val-loss is uninformative
(training loss reaches zero long before generalization).  There, spectral gap
collapse precedes grokking by hundreds to thousands of steps, demonstrating
that the geometric structure characterized in this paper has genuine predictive
utility in regimes where standard loss monitoring fails.

\paragraph{Practical considerations.}
The SED framework provides a model-agnostic diagnostic that requires
only checkpoint differences.  Combined with JL projection ($d \sim 100$--$200$
projected coordinates per checkpoint), the storage cost is negligible even for
frontier models.  Potential applications include:
\begin{enumerate}[nosep]
\item \textbf{Distribution shift detection:} The spectral edge responds within
  $\leq 2W$ steps of a shift.
\item \textbf{Training phase identification:} The three-phase spectral pattern
  identifies when optimization transitions between qualitatively different
  regimes.
\item \textbf{Early-warning signals:} In grokking-like regimes, spectral
  collapse precedes generalization events that are invisible to val-loss
  \cite{xu2026integrability}.
\end{enumerate}

\paragraph{Limitations.}
Our experiments cover two model scales (51M and 124M) with a single
architecture family (transformer) and optimizer (AdamW).  The framework's
applicability to larger models (1B+), other architectures (CNNs, state-space
models), and other optimizers (SGD, Lion, Shampoo) remains to be tested,
though the theoretical foundations (BBP, JL) are architecture-agnostic.  The
Granger causality analysis has limited statistical power due to the small
number of checkpoints (${\sim}50$), and the $p$-values should be interpreted
cautiously.  At the GPT-2 scale studied here, the spectral edge is tightly
coupled to val-loss but does not offer practical advantage over measuring
val-loss directly; the framework's distinctive utility emerges in
grokking-like regimes where loss monitoring is uninformative.

% ══════════════════════════════════════════════════════════════════════════════
\section{Conclusion}\label{sec:conclusion}

We have introduced Spectral Edge Dynamics (SED) and used it to characterize the
signal--noise geometry of neural network training across two transformer
scales (51M and 124M parameters).

The central finding is that training trajectories appear to concentrate near a
low-dimensional backbone manifold embedded in the full parameter space, with
stochastic gradients producing predominantly transverse noise.  The spectral
edge identifies the evolving boundary of this manifold: the signal rank $k^*$
adjusts automatically with task complexity ($k^* = 2$ at 51M, $k^* = 3$ at
124M), the boundary follows a universal three-phase pattern (rise, plateau,
collapse) across random seeds, and per-layer decomposition reveals uniform
spectral structure across all layer groups.  The directional Granger coupling
between the spectral edge and validation loss reverses with window size
($\abs{r} = 0.87$ at 124M, $p = 0.002$ at $W = 20$), reflecting a
timescale-dependent relationship between trajectory geometry and optimization
progress.

The framework is practical: JL random projection preserves the spectral gap
within 5--7\% at $d = 10W$, reducing storage to $O(W)$ scalars per checkpoint
independent of model size.  In companion work on algorithmic tasks
\cite{xu2026integrability, xu2026earlywarning, xu2026multitask}, the same
geometric structure provides early-warning signals of grokking---demonstrating
that the spectral edge captures optimization geometry that is invisible to
standard loss monitoring.

If the backbone manifold picture is approximately correct, it raises several
natural mathematical questions: what determines the dimension of the backbone
(the number of active Hessian modes, the rank of the Fisher information)?
When and why does the signal--noise boundary become unstable, triggering
collapse?  A companion paper will develop the theoretical foundations,
connecting BBP phase transitions, Benaych-Georges--Nadakuditi eigenvector
alignment, and Davis--Kahan subspace perturbation theory to the loss--gap
coupling characterized here.

% ══════════════════════════════════════════════════════════════════════════════
\bibliographystyle{plain}

\begin{thebibliography}{99}

\bibitem{baik2005phase}
J.~Baik, G.~Ben~Arous, and S.~P\'ech\'e.
\newblock Phase transition of the largest eigenvalue for nonnull complex sample
  covariance matrices.
\newblock {\em Annals of Probability}, 33(5):1643--1697, 2005.

\bibitem{bgn2011}
F.~Benaych-Georges and R.~R.~Nadakuditi.
\newblock The eigenvalues and eigenvectors of finite, low rank perturbations of
  large random matrices.
\newblock {\em Advances in Mathematics}, 227(1):494--521, 2011.

\bibitem{bgn2012}
F.~Benaych-Georges and R.~R.~Nadakuditi.
\newblock The singular values and vectors of low rank perturbations of large
  rectangular random matrices.
\newblock {\em Journal of Multivariate Analysis}, 111:120--135, 2012.

\bibitem{balsells2025sgd}
A.~Balsells~Rodas, Y.~Qu, and Y.~Polyanskiy.
\newblock From {SGD} to spectra: A spectral dynamics approach to neural network
  training.
\newblock In {\em ICML}, 2025.

\bibitem{cohen2021gradient}
J.~M.~Cohen, S.~Kaur, Y.~Li, J.~Z.~Kolter, and A.~Talwalkar.
\newblock Gradient descent on neural networks typically occurs at the edge of
  stability.
\newblock In {\em ICLR}, 2021.

\bibitem{davis1970rotation}
C.~Davis and W.~M.~Kahan.
\newblock The rotation of eigenvectors by a perturbation.\ {III}.
\newblock {\em SIAM Journal on Numerical Analysis}, 7(1):1--46, 1970.

\bibitem{eldan2023tinystories}
R.~Eldan and Y.~Li.
\newblock {TinyStories}: How small can language models be and still speak
  coherent {English}?
\newblock {\em arXiv preprint arXiv:2305.07759}, 2023.

\bibitem{ghorbani2019investigation}
B.~Ghorbani, S.~Krishnan, and Y.~Xiao.
\newblock An investigation into neural net optimization via {Hessian} eigenvalue
  density.
\newblock In {\em ICML}, 2019.

\bibitem{gurari2018gradient}
G.~Gur-Ari, D.~A.~Roberts, and E.~Dyer.
\newblock Gradient descent happens in a tiny subspace.
\newblock {\em arXiv preprint arXiv:1812.04754}, 2018.

\bibitem{johnson1984extensions}
W.~B.~Johnson and J.~Lindenstrauss.
\newblock Extensions of {L}ipschitz mappings into a {H}ilbert space.
\newblock {\em Contemporary Mathematics}, 26:189--206, 1984.

\bibitem{li2018measuring}
C.~Li, H.~Farkhoor, R.~Liu, and J.~Yosinski.
\newblock Measuring the intrinsic dimension of objective landscapes.
\newblock In {\em ICLR}, 2018.

\bibitem{lyu2023dichotomy}
K.~Lyu, J.~Jin, Z.~Li, S.~S.~Du, J.~D.~Lee, and W.~Hu.
\newblock Dichotomy of early and late phase implicit biases can provably induce
  grokking.
\newblock {\em arXiv preprint arXiv:2311.18817}, 2023.

\bibitem{martin2021implicit}
C.~H.~Martin and M.~W.~Mahoney.
\newblock Implicit self-regularization in deep neural networks: Evidence from
  random matrix theory and implications for training.
\newblock {\em JMLR}, 22(165):1--73, 2021.

\bibitem{merrill2023tale}
W.~Merrill, N.~Tsilivis, and A.~Shukla.
\newblock A tale of two circuits: Grokking as competition of sparse and dense
  subnetworks.
\newblock {\em arXiv preprint arXiv:2303.11873}, 2023.

\bibitem{nanda2023progress}
N.~Nanda, L.~Chan, T.~Lieberum, J.~Smith, and J.~Steinhardt.
\newblock Progress measures for grokking via mechanistic interpretability.
\newblock {\em arXiv preprint arXiv:2301.05217}, 2023.

\bibitem{pennington2017nonlinear}
J.~Pennington and P.~Worah.
\newblock Nonlinear random matrix theory for deep learning.
\newblock In {\em NeurIPS}, 2017.

\bibitem{pennington2018spectrum}
J.~Pennington and P.~Worah.
\newblock The spectrum of the {F}isher information matrix of a
  single-hidden-layer neural network.
\newblock In {\em NeurIPS}, 2018.

\bibitem{power2022grokking}
A.~Power, Y.~Burda, H.~Edwards, I.~Babuschkin, and V.~Misra.
\newblock Grokking: Generalization beyond overfitting on small algorithmic
  datasets.
\newblock In {\em ICLR Workshop on MATH-AI}, 2022.

\bibitem{radford2019language}
A.~Radford, J.~Wu, R.~Child, D.~Luan, D.~Amodei, and I.~Sutskever.
\newblock Language models are unsupervised multitask learners.
\newblock Technical report, OpenAI, 2019.

\bibitem{roberts2022principles}
D.~A.~Roberts, S.~Yaida, and B.~Hanin.
\newblock {\em The Principles of Deep Learning Theory}.
\newblock Cambridge University Press, 2022.

\bibitem{sagun2017empirical}
L.~Sagun, U.~Evci, V.~U.~G{\"u}ney, Y.~Dauphin, and L.~Bottou.
\newblock Empirical analysis of the {Hessian} of over-parametrized neural
  networks.
\newblock {\em arXiv preprint arXiv:1706.04454}, 2017.

\bibitem{xu2026optimizer}
Y.~Xu.
\newblock Optimizer-induced low-dimensional drift and transverse dynamics in
  transformer training.
\newblock {\em arXiv preprint arXiv:2602.23696}, 2026.

\bibitem{xu2026execution}
Y.~Xu.
\newblock Low-dimensional execution manifolds in transformer learning dynamics:
  Evidence from modular arithmetic tasks.
\newblock {\em arXiv preprint arXiv:2602.10496}, 2026.

\bibitem{xu2026integrability}
Y.~Xu.
\newblock Low-dimensional and transversely curved optimization dynamics in
  grokking.
\newblock {\em arXiv preprint arXiv:2602.16746}, 2026.

\bibitem{xu2026earlywarning}
Y.~Xu.
\newblock Early-warning signals of grokking via loss-landscape geometry.
\newblock {\em arXiv preprint arXiv:2602.16967}, 2026.

\bibitem{xu2026multitask}
Y.~Xu.
\newblock The geometry of multi-task grokking: Transverse instability,
  superposition, and weight decay phase structure.
\newblock {\em arXiv preprint arXiv:2602.18523}, 2026.

\end{thebibliography}

\end{document}